\documentclass[sigconf]{aamas}

\usepackage{balance} 
\usepackage{todonotes}

\renewcommand{\H}{\ensuremath{\mathcal{H}}}
\newcommand{\E}{\ensuremath{\mathcal{E}}}

\newcommand{\Pro}{\mathsf{pro}}
\newcommand{\Con}{\mathsf{con}}

\newcommand{\AR}[1]{\textcolor{black}{#1}}

\theoremstyle{definition}
\newtheorem{definition}{Definition}

\setcopyright{none}
\acmConference[arXiv]{arXiv preprint}{}{}
\acmDOI{}
\acmPrice{}
\acmISBN{}

\acmSubmissionID{<<submission id>>}

\title[AAMAS-2026 Formatting Instructions]{Towards an Argumentative Foundation for Evaluative AI}

\author{Xiang Yin}
\affiliation{
  \institution{Imperial College London}
  \city{London}
  \country{United Kingdom}}
\email{xy620@ic.ac.uk}

\author{Tim Miller}
\affiliation{
  \institution{The University of Queensland}
  \city{Brisbane}
  \country{Australia}}
\email{timothy.miller@uq.edu.au}

\author{Nico Potyka}
\affiliation{
  \institution{Cardiff University}
  \city{Cardiff}
  \country{United Kingdom}}
\email{PotykaN@cardiff.ac.uk}

\author{Antonio Rago}
\affiliation{
  \institution{King's College London}
  \city{London}
  \country{United Kingdom}}
\email{antonio.rago@kcl.ac.uk}

\author{Francesca Toni}
\affiliation{
  \institution{Imperial College London}
  \city{London}
  \country{United Kingdom}}
\email{ft@ic.ac.uk}

\begin{abstract}
Evaluative AI (EAI) has been recently proposed as a way to support human decision-making, not by producing a single recommendation, but by presenting competing hypotheses together with evidence for and against each. In this position paper, we advocate (computational) argumentation as a most suitable paradigm to provide a formal, computable foundation for forms of EAI that are explainable and contestable. Argumentation can also naturally pave the way to a multi-agent vision for EAI in which diverse evaluative models, derived from different sources and reasoning styles, can interact and jointly deliberate on hypotheses and evidence. Overall, this position paper sets the ground for a long-term research agenda towards distributed and human-centred EAI systems.
\end{abstract}


\newcommand{\BibTeX}{\rm B\kern-.05em{\sc i\kern-.025em b}\kern-.08em\TeX}

\begin{document}


\pagestyle{fancy}
\fancyhead{}


\maketitle 


\section{Introduction}
\label{sec:introduction}

Explainable AI (XAI) aims to enhance the transparency and trustworthiness of AI systems by elucidating their decision-making processes~\cite{adadi2018peeking}, a capability that is essential in high-stakes domains such as healthcare, the judiciary, and finance.
\emph{Evaluative AI (EAI)} \cite{EAI_Tim_Miller} has been recently proposed as a new form of XAI to support human decision-making 
in a \emph{hypothesis-driven} rather than 
\emph{recommendation-driven} fashion. 
The latter 
is a widely used paradigm in XAI: an AI model first produces an output, and XAI methods subsequently explain or justify it (e.g., \cite{lundberg2017unified,ribeiro2016should,wachter2017counterfactual}). Yet, recent studies show that such ``recommend-and-explain’’ paradigm can induce cognitive fixation, leading users to either over-rely on or dismiss AI recommendations without sufficient deliberation \cite{buccinca2021trust,gajos2022people,sivaraman2023ignore}.
To address these issues, instead of directly presenting a recommendation, EAI offers multiple plausible \emph{hypotheses} together with structured \emph{evidence} for and against each. Thus, by ensuring that the human actively evaluates and deliberates between competing hypotheses, EAI preserves human agency in decision-making. 
This, in turn, strengthens the human-in-the-loop paradigm and repositions AI systems from decision-makers to facilitators.

In this position paper we set the ground for 
a long-term research agenda 
towards distributed 
and human-centred EAI systems with clear formal and algorithmic backings.
Specifically, we 
adopt a
formal understanding of the EAI problem as a \emph{ranking-based} problem over all potential hypotheses, given the pro and con evidence associated with each, as was suggested as future work by \citet{EAI_Tim_Miller}.
For example, in healthcare, given a patient's symptoms (evidence), 
an EAI-driven system may generate a ranked list of possible diagnoses (hypotheses) to assist
practitioners in their decision-making process.
We see 
a ranking over the hypotheses 
not as a final decision but, rather, as reflecting  
the ``preferences'' of the EAI-driven system proposing it,
given the evidence.
Then, our main position is that 
 (computational) argumentation,  
 particularly through the use of weighted Quantitative Bipolar Argumentation Frameworks (wQBAFs) \cite{mossakowski2018modular,chi2021optimized,potyka2021interpreting},
 is a most suitable
 paradigm 
 to provide a formal and computational
 foundation for ranking-based EAI,
with advantages over the originally proposed Weight of Evidence (WoE) method \cite{Tim_WoE} in terms of reasoning with structured hypotheses and  evidence, 
explainability~\cite{vcyras2021argumentative,vassiliades2021argumentation}, human engagement (in particular as concerns the possibility for humans to contest~\cite{leofante2024contestable} specifically evidence, hypotheses, and ultimately rankings), and finally
distributed variants of EAI where diverse 
\AR{agents}, derived from different sources and reasoning styles, can
interact and jointly deliberate on hypotheses and evidence.

\begin{figure}[t]
    \centering
    \includegraphics[width=1.0\linewidth]{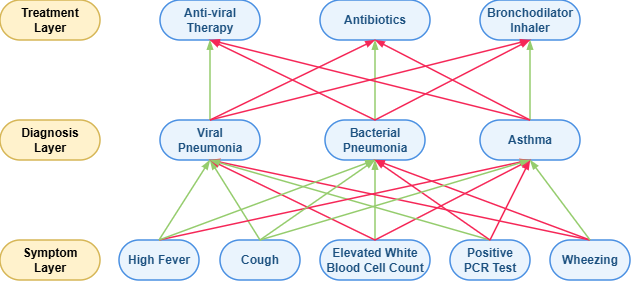}
    \caption{
    Skeleton of an argumentative solution to an example ranking-based EAI problem in healthcare. Blue nodes denote arguments; green/red edges indicate, resp., support/attack relations. If, in the corresponding wQBAF, all arguments have an initial weight of 0.5 (the neutral value in [0,1]) and the edges all have the same edge weight of 1 (the top value in [0,1]) , then the O-QuAD semantics \cite{chi2021optimized} yields the ranking $\texttt{Anti-viral} 
    = \texttt{Antibiotics} \succ \texttt{Bronchodilator}$ with corresponding strengths 0.40, 0.40, 0.31.}
    \label{fig_qbaf}
\end{figure}

\paragraph{\textbf{Motivating Illustration}.}
Figure \ref{fig_qbaf}
depicts  the skeleton of an argumentative solution to the ranking-based EAI problem in a healthcare scenario\footnote{This is a simplified example used solely for illustration and does not reflect clinically validated relations.}.
This amounts to  
\emph{arguments}, organised hierarchically across three layers via \emph{attack and support relations}.
The arguments in the `symptom layer' represent observable evidence (e.g., high fever, cough), in the `treatment layer' represent possible clinical decisions (e.g., anti-viral therapy, antibiotics), and in the intermediate `diagnosis layer' (e.g., viral/bacterial pneumonia, asthma) play a dual role: as hypotheses with respect to the evidence below, but also as evidence for the hypotheses above.\footnote{Note that our argumentative solution is not restricted to 3-layer structure, and can accommodate any structure.}
The 
support and attack relations reflect dependencies between hypotheses and evidence.
A wQBAF then augments this  skeleton by associating to each argument
 an \emph{initial weight}, which may represent their importance or the  confidence level in its validity,
 and to each edge a \emph{relation weight}, which  may capture the influence of the edge towards the affected argument.
 Further, quantitative \emph{evaluation methods}
(e.g., \cite{chi2021optimized,potyka2021interpreting}
can be used to determine the \emph{(final) strength} of each argument, recursively based on its own initial weight and the combined strengths of its supporters and attackers,
enabling 
conflict resolution with the available information~\cite{vcyras2021argumentative,potyka2021interpreting,ayoobi2023sparx,potyka2023explaining}.
Finally,
from the 
 arguments' strength, we can obtain a ranked list of the hypotheses in the treatment layer.
The wQBAF and associated strengths are interpretable and can be used for explanation of the ranking and/or the validity of any hypothesis or evidence, in the spirit of~\cite{vcyras2021argumentative,vassiliades2021argumentation}, in a variety of formats, e.g., \cite{kampik2022explaining,AAE_ECAI,amgoud2017measuring,YIN_RAE_IJCAI,AAEsRAEs-JAIR24,Caren_2025impactmeasure,Tim_set_contribution}  
Moreover, should a human disagree with any component of the QBAF and/or the ranking, they can contest it, in the spirit of \cite{yin2025contestability}, e.g. to disagree with the initial weight of some evidence and/or add additional pro and con evidence or hypotheses.
 This ability to contest promotes human-in-the-loop decision-making, which lies at the heart of EAI.
Finally, the wQBAF may present the opinion of an agent and could be integrated with opinions by other agents, e.g., with argumentative exchanges as in \cite{rago2023interactive}, or result from the aggregation of QBAFs from different agents, e.g., as in \cite{merging_argumentation}.

\paragraph{\textbf{Contributions}.}
The vision of this position paper is threefold.
\begin{itemize}
\item 
 EAI can be formally understood as a ranking-based problem (Section~\ref{sec:main}). 
\item 
wQBAFs from the field of (computational) argumentation  provide a principled paradigm for realising ranking-based EAI in an explainable and contestable manner suitable for human-centred EAI systems (Section~\ref{sec:qbafs}).
\item 
wQBAFs can empower agent-level EAI within a multi-agent vision for EAI, in which diverse evaluative models interact and jointly deliberate (Section~\ref{sec:mas}).
\end{itemize}

\section{Related Work}
\label{sec:related_work}

\paragraph{\textbf{Evaluative AI (EAI)}}
EAI is a recent paradigm \cite{EAI_Tim_Miller}, and its practical realisation is still in the early stages of investigation.
Only a handful of approaches have been proposed so far to realise EAI, notably~\cite{Tim_WoE,ermellino2024approach}: 
\citet{Tim_WoE} adopt Weight of Evidence (WoE) \cite{melis2021human}  to quantify both the direction (supporting or opposing) and the magnitude of the impact that each piece of evidence has on candidate hypotheses; and \citet{ermellino2024approach} utilise large language models (LLMs) to generate pro and con evidence in a conversational style. 
In contrast to these approaches, our envisaged ranking-based argumentative approach to EAI allows richer structures of dependencies between hypotheses and evidence, e.g., hypotheses may influence other hypotheses.
It also allows (i) to separate the stance of evidence, (ii) to determine the rating and ranking of hypotheses via existing evaluation methods from wQBAFs, benefiting from an extensive literature on their formal and practical properties \cite{baroni2018many,yin2025contestability}
and (iii) to derive faithful explanations for the rankings from the evaluations,  while also (iv) mitigating against hallucination risks when evidence and hypotheses are drawn from LLMs with the help of the contestable nature of argumentative solutions~\cite{argllms}.


\paragraph{\textbf{Argumentative XAI}} Argumentation has emerged as an influential paradigm for supporting XAI 
with structured, transparent, and human-aligned reasoning processes (see recent surveys \cite{vcyras2021argumentative,vassiliades2021argumentation}). 
Our EAI vision can be seen as falling under the category of \emph{intrinsic} argumentative explanations~\cite{vcyras2021argumentative} in which the underlying model is already argumentative (in our vision, a wQBAF).
We can then leverage on a variety of explanation formats 
for explaining rankings between hypotheses,  e.g. 
based on assessing the most influential evidence for or against various ranked hypotheses, in the spirit of
\emph{argument/relation attribution explanations}~\cite{kampik2022explaining,AAE_ECAI,YIN_RAE_IJCAI,AAEsRAEs-JAIR24,amgoud2017measuring,
Caren_2025impactmeasure,Tim_set_contribution}.
For instance, the top-ranked hypothesis \texttt{Anti-viral} (Figure~\ref{fig_qbaf}) attains a strength of 0.40, where \texttt{High Fever}, \texttt{Cough} and \texttt{Positive PCR test} are computed as the most influential symptoms following the approach of~\cite{AAE_ECAI}.
An important aspect of explanations drawn from intrinsically argumentative solutions is that they are by definition \emph{faithful}~\cite{argllms}.
We envisage that this is an important endorsement for our vision for trustworthy EAI.

\paragraph{\textbf{Contestability}.} The need for AI to be contestable in general is widely acknowledged~\cite{contestableAI-alfrink}, and argumentation is advocated by several as a most suitable formalism to achieve principled forms of contestable AI  \cite{leofante2024contestable,aamas25contestableAI}. 
Contestable argumentative solutions have been proposed, e.g., 
when QBAFs are generated by LLMs~\cite{argllms} and
with the help of 
\emph{counterfactual explanations}~\cite{yin2024qarg,kampik2024change,yin2025contestability}.
The latter
indicate how to modify the weights of arguments or relations to change the strength of a hypothesis to a desired one, which allow users to interact with and challenge EAI systems built argumentatively.
The contestability naturally afforded 
by argumentative solution is a strong 
motivation for our vision of ranking-based argumentative EAI that is amenable to human consumption.

\paragraph{\textbf{Multi-Agent Argumentation}.}
Several works envisage argumentation as the basis for multi-agent interaction and deliberation, and can serve as a starting point for 
a multi-agent vision of argumentative EAI.
Existing approaches 
are based on static 
aggregation of argumentation frameworks from various agents, e.g., 
abstract argumentation frameworks in \cite{merging_argumentation} and
bipolar argumentation frameworks in 
\cite{DBLP:journals/aamas/DickieLBRT25},
or 
on dynamic, dialogue-based combinations,
e.g., of weighted abstract argumentation frameworks in \cite{DBLP:conf/atal/TarleBM22} and of QBAFs in \cite{rago2023interactive}.
The latter could be especially useful as a starting point to support our vision of wQBAF-based EAI.

\section{The Ranking-based EAI Problem}
\label{sec:main}

Let $\mathcal{H}$ be a finite, non-empty set of \emph{hypotheses}, and $\mathcal{E}$ be a finite set of (pieces of) \emph{evidence}.
%
%
%
Hypotheses and evidence can be associated with an \emph{initial weight} that capture how plausible 
they are before considering any 
dependencies amongst them.
\begin{definition}[Initial Weight]
\label{def_initial_weight}
The function $\tau: (\E \cup \H) \rightarrow [0,1]$ maps each $x \in (\E \cup \H)$ to its \emph{initial weight} $\tau(x)$.
\end{definition}
\noindent

In the motivating illustration, the initial weight of a symptom (e.g., \texttt{cough}) or of a medical condition (e.g., \texttt{asthma}) may capture its perceived severity for a given patient. For instance, a \texttt{higher fever} corresponds to a higher initial weight, reflecting the greater severity of the symptom.

(Positive and negative) dependencies within 
hypotheses and evidence can be modelled in terms of 
(pro and con, resp.) relations.

\begin{definition}[
Pro/Con Relations]
\label{def:procon}
    $\Pro, \Con \subseteq 
    (\E \times (\E \cup \H)) \cup 
    (\H\times \H)$ are disjoint binary relations,  referred to, resp., as the \emph{pro} and \emph{con relations}.
    For all $x,y \in (\E \cup \H)$
    ,
    $x$ is called \emph{pro/con evidence or hypothesis} for $y$ iff $(x,y) \in \Pro$/$(x,y) \in \Con$, resp.
\end{definition}

Note that here
we consider, besides direct relations from evidence to hypotheses ($\E \times \H$) as in \cite{EAI_Tim_Miller}, also relations among evidence ($\E \times \E$) and among hypotheses ($\H \times \H$).  In this way, evidence may also attack or support other evidence. 
This extension allows multi-step reasoning and evaluation.

To quantify how strongly a piece of evidence affects the plausibility of another element (hypothesis or evidence), we use 
the notion of \emph{relation weight}.
\begin{definition}[Relation Weight]
\label{def_weight_relation}
The function $w:  (\Pro \cup \Con) \rightarrow [0,1]$ maps each $r \in (\Pro \cup \Con)$ to its \emph{relation weight} $w(r)$.
\end{definition}

In the motivating illustration, the relation weight for an evidence-hypothesis pair (e.g., \texttt{(cough,asthma)}) represents the strength of the influence from the symptom \texttt{cough} to the diagnosis \texttt{asthma}. A higher weight indicates a stronger influence.

We next formally define EAI as a \emph{ranking-based problem}.
\begin{definition}[The Ranking-based EAI Problem]
    Given $E=\left\langle \E, \H,\right.$
    $\left.\!\Pro, \Con, \tau, w \right\rangle$,
    the \emph{ranking-based EAI problem} is to determine a total preorder $\succcurlyeq_{E}$ on $\mathcal{H}
    $.~\footnote{A total preorder is a reflexive, transitive and total relation.}
    \end{definition}
    Intuitively,
$\succcurlyeq_{E}$
    represents the relative \emph{plausibility} of hypotheses given the weighted pro and con relations, and the initial weights of evidence and hypotheses.
For any $h_1, h_2 \in \mathcal{H}$, $h_1 \succcurlyeq_{E} h_2$ implies that $h_2$ is not more plausible than $h_1$ given $E$.

Formalising EAI as a ranking-based problem offers 
two key advantages.
First, ranking naturally accommodates multiple hypotheses rather than forcing a single recommendation, which aligns directly with the core philosophy of EAI.
Second, 
ranking offers a concrete computational framework, both for generating rankings (e.g., \cite{herbrich2000large, burges2005learning}) and for evaluating them (e.g., Kendall’s $\tau$ \cite{ kendall1938new}), which transforms EAI from a conceptual vision into a practicable research agenda.
We discuss next how to use argumentation 
to identify solutions to the ranking-based EAI problem.





\section{An Argumentative Solution}
\label{sec:qbafs}
In this section, we advocate that 
argumentation can 
lead to the identification of solutions to the ranking-based EAI problem with desirable properties.
Specifically, we envisage the use of
\emph{weighted Quantitative Bipolar Argumentation
Frameworks (wQBAFs)}\cite{mossakowski2018modular,chi2021optimized,potyka2021interpreting}, which are
\label{def_QBAF}
 quintuples $\mathcal{Q}=\left\langle\mathcal{A}, \mathcal{R}^{-}, \mathcal{R}^{+}, \tau, w  \right\rangle$ where $\mathcal{A}$ is a finite set of \emph{arguments}, $\mathcal{R}^{-}, \mathcal{R}^{+} \subseteq \mathcal{A} \times \mathcal{A}$ are disjoint binary relations called \emph{attack} and \emph{support}, $\tau: \mathcal{A} \rightarrow  [0,1]$ is a \emph{base score function}, and $w: \mathcal{{R}^{-}\cup{R}^{+}} \rightarrow [0,1]$ is an \emph{edge weight function}.
These 
wQBAFs can be used to represent EAI problems.
\begin{definition}[wQBAF Representation for 
EAI Problems]
The  \emph{wQBAF representation of}
$E=\left\langle \E, \H,\right.$
    $\left.\!\Pro, \Con, \tau, w \right\rangle$
 is the wQBAF $\mathcal{Q} = \langle \mathcal{A}, \mathcal{R}^{+}, \mathcal{R}^{-}, \tau, w \rangle$ such that:
    \begin{itemize}
        \item $\mathcal{A} = \E \cup \H$;
        \item $\mathcal{R}^{-} =\Con 
        $; \item 
        $ \mathcal{R}^{+} =\Pro 
        $
        .
    \end{itemize}
\end{definition}

We define $\mathcal{A}$ as the set $\H \cup \E$, so that debate can take place about all evidence and hypotheses.
As for the other components, we use those of the EAI problem directly.



Once we map the EAI problem into an argumentative representation,
identifying solutions to the 
ranking-based EAI problem requires 
two steps.
First, we compute the strengths of arguments (both evidence and hypotheses) using quantitative evaluation methods, which update the initial weights of arguments by taking support and attack relations into account (see, e.g., \cite{chi2021optimized,potyka2021interpreting}).\footnote{Existing ranking-based semantics~\cite{amgoud2013ranking} are defined for abstract argumentation frameworks and cannot be directly adapted to our quantitative setting.
}
Second, we derive a ranking over hypotheses by comparing their strengths, as illustrated in the Introduction.
The 
wQBAF solution to 
the ranking-based EAI problem is naturally explainable: qualitatively, the graphical structure shows the 
relationship between different arguments and users can see the reasoning path from a piece of evidence to a hypothesis; quantitatively, the final strength of each hypothesis is explainable via many possible explanation method for QBAFs, such as attribution method \cite{AAE_ECAI,YIN_RAE_IJCAI}
and counterfactual explanations \cite{yin2024qarg,yin2025contestability},
affording also contestability by humans who
 can modify the weights of arguments or relations to change the strength of a hypothesis to a desired one
 .
Our envisaged argumentative solution also leads naturally to the satisfaction of a number of desirable principles for EAI, introduced. 
These principles serve both as design guidelines for selecting or developing ranking methods and as criteria for evaluating their behaviour.

\paragraph{\textbf{Point-wise Principles}}
Point-wise principles consider the ranking of individual hypotheses.

\textbf{Principle 1 (Monotonicity).} Monotonicity is a fundamental guarantee 
to ensure 
that rankings remain aligned with rational intuition and expectation. Monotonicity requires that the relative ranking of an individual hypothesis should change in a way consistent with the direction of change in the underlying factors, such as the introduction of new evidence or updates to the weights of existing evidence. For instance, adding additional pro evidence for a hypothesis should never cause its rank to fall; and increasing the initial weight of its piece of pro evidence should not lower its position relative to others. 

Many widely used evaluation methods for wQBAFs
(e.g., O-QuAD \cite{chi2021optimized}, MLP-based semantics \cite{potyka2021interpreting}, and edge-weighted variants of REB \cite{amgoud2018evaluation} and QE \cite{Potyka18}) are known to satisfy Principle 1 \cite{yin2025contestability}.

\textbf{Principle 2 (Balance).}
Balance ensures that a ranking remains unchanged when the pro and con evidence are balanced. In particular, any modification of the evidence that preserves the equality between the overall strength of pro and con evidence should not affect the relative ranking of hypotheses. This includes, for instance, redistributing weights among existing pro and con evidence or introducing additional evidence, provided that the total pro and total con influence remain equal and therefore cancel each other out.

The previously mentioned semantics also satisfy Principle 2.

\paragraph{\textbf{Pair-wise Principles}}
Pair-wise principles consider the ranking comparison of two hypotheses.

\textbf{Principle 3 (Equivalence).} Equivalence states that two hypotheses must be ranked equally if they share the same set of pro and con evidence with the same initial weight. This principle is fundamental for ranking fairness, ensuring symmetric treatment for evidentially symmetric cases. Beyond fairness, it is also a prerequisite for rational and intuitive ranking method behaviour.

The previously mentioned semantics also satisfy Principle 3.

\textbf{Principle 4 (Dominance).} The Dominance principle establishes a preference in the ranking under controlled conditions. For example, a hypothesis must be ranked no lower than another if, given an equivalent set of con evidence against both, either (1) its set of pro evidence is a superset of the other's, meaning that more pro evidence ranking higher; or (2) the aggregated weight of its pro evidence is greater. This principle is fundamental because it ensures the ranking objectively rewards hypotheses with superior pro evidence (either in number or in strengths), providing a clear and fair rationale for comparative assessment.

The previously mentioned semantics also satisfy Principle 4.

\paragraph{\textbf{General Principles}}
While pointwise and pairwise principles ensure local rationality, 
general principles govern the global behaviour of rankings. We envision two key principles at this level.

\textbf{Principle 5 (Robustness).} Robustness requires that the overall ranking exhibits stability when minor perturbations to the underlying evidence or its associated weights occur. Specifically, small variations in the evidence should not induce large-scale or counter-intuitive reversals in the resulting ranking order. This principle is critical for ensuring the reliability and practical dependability of the EAI rankings, as users must be able to trust that its evaluations are not brittle or overly sensitive to noise.

It is unclear whether existing evaluation methods for wQBAFs ensure
robustness at the ranking level; if not, new robustness-oriented
ranking semantics may be needed.

\textbf{Principle 6 (Explainability).} Explainability requires that the ranking is generated through a mechanism whose logic is transparent and whose outcomes can be clearly explained. This entails providing both local explanations (e.g., why one hypothesis is ranked above another) and a global rationale for the overall ordering. This principle is fundamental to transforming the ranking from an opaque output to an understandable result, thereby fostering user comprehension and trust.

As we already argued earlier, 
a large array of 
argumentative explanations drawn from wQBAFs can support this principle.

\textbf{Principle 7 (Contestability).} Contestability ensures that the EAI ranking can be challenged by users. This includes the ability to question not only the veracity of individual evidence, or their initial weights assigned, but also the final ranking itself. By explicitly supporting such challenges, this principle reaffirms the role of EAI as a deliberative partner in a human-AI team, empowering users to engage with and refine the system's reasoning process critically.

Again, as already argued earlier, 
wQBAFs show promise to support this principle.

\section{Multi-agent EAI via Argumentation}
\label{sec:mas}
Real-world evaluative processes rarely stem from a single viewpoint: they integrate heterogeneous information sources (from human experts to domain tools and even LLM-generated evidence), diverse ways of identifying argumentative relations, and different reasoning mechanisms induced by alternative gradual semantics. 
In the illustration,
different agents may contribute the parts of the wQBAF relative to the different hypotheses in the treatment layer, reflecting different medical expertise.
A multi-agent perspective therefore not only mitigates individual bias but also enables richer, more complementary evidence and reasoning styles, opening the door to more robust and trustworthy evaluations. 
Viewing each wQBAF as an autonomous argumentative agent suggests a broader research vision: agents may engage in argumentative communication, exchanging arguments or relations to surface new evidence, reconcile inconsistencies, or converge through principled exchange protocols \cite{rago2023interactive}. 
Also, multiple wQBAFs may be fused into a group-level evaluation via semantic alignment of arguments, clustering of evidence structures \cite{gorur2025retrieval}, or ensemble-style aggregation over hypothesis rankings \cite{ganaie2022ensemble}. This multi-agent outlook positions argumentation-based EAI as the foundation for a distributed, resilient, and genuinely deliberative form of EAI, where interacting argumentative agents 
together construct evaluations that have the potential not only of being less biased and more robust, but also 
explainable and contestable.

\section{Discussion}
 The primary impact of this position paper lies in transitioning EAI from a conceptual proposal to a tractable research agenda by establishing a formal, ranking- and argumentation-based foundation. 
 Furthermore, it inherently paves the way towards 
 explainability, as the argumentative structure makes the entire evaluation process explicit and auditable. Finally, this explicit reasoning process empowers contestability, transforming the AI from a static recommender into a dynamic deliberative aider that users can meaningfully challenge and refine, thereby fostering a new paradigm of collaborative and trustworthy human-in-the-loop decision-making.


Some potential limitations may shape the scope of our vision. First, our framework assumes that wQBAFs, including arguments, relations, and associated credibility score, are already correctly constructed, whereas in practice the reliable extraction of argumentative structure remains a core bottleneck; advances in LLM-based argument mining may help alleviate this challenge \cite{gorur2025can,gorur2025retrieval}. Building on this, even when a wQBAF is successfully constructed, the openness required for contestability may introduce a second limitation: users may strategically or unintentionally provide misleading or low-quality evidence, creating risks of manipulation; mechanisms such as permissioned contribution and trust-aware weighting may offer partial safeguards. Finally, beyond individual agents, multi-agent EAI systems may naturally exhibit deep and sometimes irreducible disagreement \cite{wu2025hidden}. Rather than enforcing consensus, this limitation suggests the need for managing such pluralism, supporting structured disagreement \cite{liang2024encouraging}, and helping users navigate conflicting evaluations.

\bibliographystyle{ACM-Reference-Format}
\bibliography{sample}

\end{document}